\documentclass[10pt,journal]{IEEEtran}
\usepackage[utf8]{inputenc}

\newcommand{\rev}[1]{#1}

\ifCLASSOPTIONcompsoc
 \usepackage[nocompress]{cite}
\else
 \usepackage{cite}
\fi
\usepackage{comment}
\usepackage{amsmath}
\usepackage{amssymb}
\usepackage{png-math}
\usepackage{png-tikz}
\usepackage{graphicx}
\usepackage{doi}

\usepackage{hyperref} 

\hypersetup{
   colorlinks,
   menucolor=black,
   linkcolor=black,
   citecolor=black,
   urlcolor=blue
}

\newcommand{\weight}[0]{\omega}
\newcommand{\bias}[0]{\upsilon}
\newcommand{\Weight}[0]{\Vec{\weight}}

\newcommand{\Alpha}[0]{\Vec{\alpha}}

\newcommand{\param}[0]{\theta}
\newcommand{\Param}[0]{\Vec{\param}}

\newcommand{\class}[0]{c}

\newcommand{\citep}[1]{\cite{#1}}

\newcommand{\W}[0]{\Vec{\omega}}
\newcommand{\U}[0]{\Vec{\upsilon}}
\newcommand{\V}[0]{\nu}
\newcommand{\Hidden}[0]{\Vec{h}}
\newcommand{\CEC}[0]{c}

\newcommand{\cont}[0]{z}
\newcommand{\back}[0]{s}
\newcommand{\cvar}[0]{\zeta}
\newcommand{\feat}[0]{\phi}
\newcommand{\Feat}[0]{\Vec{\feat}}
\newcommand{\obs}[0]{x}
\newcommand{\Obs}[0]{\Vec{\obs}}
\newcommand{\OBS}[0]{\Vec{X}}
\newcommand{\hid}[0]{h}
\newcommand{\Hid}[0]{\Vec{\hid}}
\newcommand{\HID}[0]{\Vec{H}}

\newcommand{\rlvn}[0]{r}
\newcommand{\rvar}[0]{\rho}

\usetikzlibrary{fit}
\usetikzlibrary{positioning}
\usetikzlibrary{calc}
\tikzstyle{op} = [circle,draw=black,inner sep=0ex]

\begin{document}
%
\title{A Bayesian Approach to Recurrence\\ in Neural Networks}

%
%
%
%

\author{%
  Philip~N.~Garner,\ 
  Sibo~Tong
  \thanks{\copyright~2020 IEEE.  Personal use of this material is permitted.  Permission from IEEE must be obtained for all other uses, in any current or future media, including reprinting/republishing this material for advertising or promotional purposes, creating new collective works, for resale or redistribution to servers or lists, or reuse of any copyrighted component of this work in other works.}
  \thanks{\href{https://dx.doi.org/10.1109/TPAMI.2020.2976978}{doi:10.1109/TPAMI.2020.2976978}}
}

\IEEEtitleabstractindextext{%
  \begin{abstract}
    We begin by reiterating that common neural network activation functions have
simple Bayesian origins.  In this spirit, we go on to show that Bayes's theorem
also implies a simple recurrence relation; this leads to a Bayesian recurrent
unit with a prescribed feedback formulation.  We show that introduction of a
context indicator leads to a variable feedback that is similar to the forget
mechanism in conventional recurrent units.  A similar approach leads to a
probabilistic input gate.  The Bayesian formulation leads naturally to the two
pass algorithm of the Kalman smoother or forward-backward algorithm, meaning
that inference naturally depends upon future inputs as well as past ones.
Experiments on speech recognition confirm that the resulting architecture can
perform as well as a bidirectional recurrent network with the same number of
parameters as a unidirectional one.  Further, when configured explicitly
bidirectionally, the architecture can exceed the performance of a conventional
bidirectional recurrence.


  \end{abstract}
}

\maketitle

\IEEEdisplaynontitleabstractindextext

%
\IEEEpeerreviewmaketitle


%
%
%
%


\section{Introduction}

\IEEEPARstart{I}{n} signal processing and statistical pattern recognition,
recurrent models have been ubiquitous for some time.  They are perhaps
exemplified by two cases: the state space filter of Kalman \cite{Kalman1960,
  Scharf1991} is appropriate for continuous states; the hidden Markov model
(HMM) \cite{Baum1966, Bahl1983} for discrete states.  Both of these approaches
can be characterised as being statistically rigorous; each has a
forward-backward training procedure that arises from a statistical estimation
formulation.

\begin{figure}[htb]
  \centering
  \resizebox{0.9\columnwidth}{!}{\begin{tikzpicture}[every fit/.style={rectangle,draw,inner sep=1.5ex}]

  \node (Linear) [c] {$+$};
  
  \node (fGate) [op,above=of Linear] {$\times$};
  \draw[a] (Linear) to [bend right=60] node [right] {$c_{t-1}$} (fGate.east);
  \draw[a] (fGate.west) to [bend right=60] (Linear);
  \node [node distance=1ex,left=of fGate] {\small Forget};

  \node (iGate) [op,left=of Linear] {$\times$};
  \draw[a] (iGate) to node [above left] {$\tilde{c}_t$} (Linear);
  \node [node distance=1ex,above=of iGate] {\small Input};

  \node (iPsi) [c,left=of iGate] {$\psi_i$};
  \draw[a] (iPsi) to (iGate);
  \node (iScalar) [left=of iPsi,anchor=east] {$\W\Scalar_t$};
  \node (iHidden) [above=of iScalar] {$\U\Hidden_{t-1}$};
  \draw[a] (iScalar) to (iPsi);
  \draw[a] (iHidden) to (iPsi);

  \node (oPsi) [c,right=of Linear] {$\psi_o$};
  \draw[a] (Linear) to node [above] {$c_t$} (oPsi);
  \node (oGate) [op,right=of oPsi] {$\times$};
  \draw[a] (oPsi) to (oGate);

  \node (peep) [node distance=10ex,below right of=Linear] {$c_t$};
  \draw[a] (Linear) to [bend left=30] (peep);

  \node (output) [right=of oGate] {$h_t$};
  \draw[a] (oGate) to (output);
  \node (outlab) [node distance=1ex,above=of oGate] {\small Output};

  \node (oSig) [c,below=of oGate] {$\sigma$};
  \node [above right] at (oSig.north) {$o_t$};
  \draw[a] (oSig) to (oGate);
  \node (o2) [below=of oSig] {$\U_o\Hidden_{t-1}$};
  \node (o1) [node distance=1ex,left=of o2] {$\W_o\Scalar_t$};
  \node (o3) [node distance=1ex,right=of o2] {$\V_o\CEC_t$};
  \draw[a] (o1) to (oSig);
  \draw[a] (o2) to (oSig);
  \draw[a] (o3) to (oSig);

  \node (fSig) [c,above=of fGate] {$\sigma$};
  \node [below right] at (fSig.south) {$f_t$};
  \draw[a] (fSig) to (fGate);
  \node (f2) [above=of fSig] {$\U_f\Hidden_{t-1}$};
  \node (f1) [node distance=1ex,left=of f2] {$\W_f\Scalar_t$};
  \node (f3) [node distance=1ex,right=of f2] {$\V_f\CEC_{t-1}$};
  \draw[a] (f1) to (fSig);
  \draw[a] (f2) to (fSig);
  \draw[a] (f3) to (fSig);

  \node (iSig) [c,below=of iGate] {$\sigma$};
  \node [above right] at (iSig.north) {$i_t$};
  \draw[a] (iSig) to (iGate);
  \node (i2) [below=of iSig] {$\U_i\Hidden_{t-1}$};
  \node (i1) [node distance=1ex,left=of i2] {$\W_i\Scalar_t$};
  \node (i3) [node distance=1ex,right=of i2] {$\V_i\CEC_{t-1}$};
  \draw[a] (i1) to (iSig);
  \draw[a] (i2) to (iSig);
  \draw[a] (i3) to (iSig);

  \node[fit=(iPsi) (iGate) (Linear) (fGate) (oPsi) (oGate) (outlab)] (box) {};
  \node[below left] at (box.north east) {LSTM cell};
\end{tikzpicture}

}
  \caption{The long short term memory of \cite{Hochreiter1997}.
    Non-linearities $\psi$ are taken to be $\tanh$.}
  \label{fig:LSTM}
\end{figure}
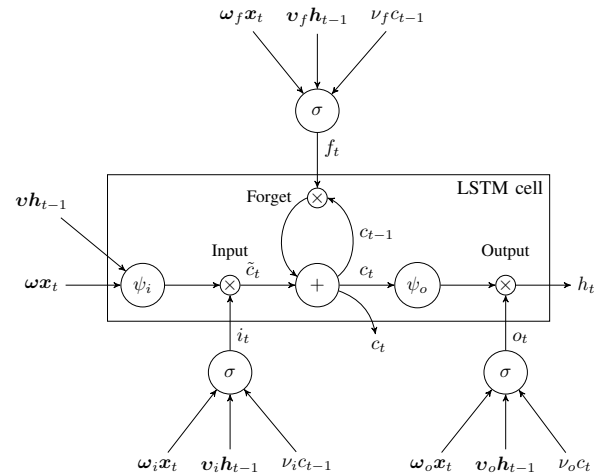

Recurrence is also important in modern deep learning.  The foundations were
laid shortly after the introduction of the multi-layer perceptron (MLP)
\cite{Rumelhart1986a, Rumelhart1986b} with the back-propagation through time
algorithm \cite{Rumelhart1986a, Williams1989}.  Such architectures can be
difficult to train; some of the difficulties were addressed by the long
short-term memory (LSTM) of Hochreiter and Schmidhuber \cite{Hochreiter1997}.
The LSTM was subsequently modified by Gers et al.~\cite{Gers2000} to include a
forget gate, and by Gers et al.~\cite{Gers2002} to include peephole connections.
The full LSTM is illustrated in figure \ref{fig:LSTM}.

\begin{figure}[htb]
  \centering
  \resizebox{0.8\columnwidth}{!}{\begin{tikzpicture}[every fit/.style={rectangle,draw,inner sep=1.5ex}]

  \node (Linear) [c] {$+$};
  
  \node (uGate) [op,above=of Linear] {$\times$};
  \draw[a] (Linear) to [bend right=60] node [right] {$h_{t-1}$} (uGate.east);
  \draw[a] (uGate.west) to [bend right=60] (Linear);
  \node (uplab) [node distance=1ex,right=of uGate] {\small Update};

  \node (iGate) [op,left=of Linear] {$\times$};
  \draw[a] (iGate) to node [below] {$\tilde{h}_t$} (Linear);

  \node (iPsi) [c,left=of iGate] {$\psi_i$};
  \draw[a] (iPsi) to (iGate);
  \node (rGate) [op,above left=of iPsi] {$\times$};
  \node (iHidden) [left of=rGate,anchor=east] {$\U\Hidden_{t-1}$};
  \node (iScalar) [node distance=12ex,left=of iPsi] {$\W\Scalar_t$};
  \draw[a] (iScalar) to (iPsi);
  \draw[a] (iHidden) to (rGate);
  \draw[a] (rGate) to (iPsi);
  \node [node distance=0.3ex,right=of rGate] {\small Reset};

  \node (output) [node distance=12ex,right=of Linear] {$h_t$};
  \draw[a] (Linear) to (output);

  \node (rSig) [c,above=of rGate] {$\sigma$};
  \node [below right] at (rSig.south) {$r_t$};
  \draw[a] (rSig) to (rGate);
  \node (rDummy) [above=of rSig] {};
  \node (r1) [node distance=1ex,left=of rDummy] {$\W_r\Scalar_t$};
  \node (r2) [node distance=1ex,right=of rDummy] {$\U_r\Hidden_{t-1}$};
  \draw[a] (r1) to (rSig);
  \draw[a] (r2) to (rSig);

  \node (uSig) [c,above=of uGate,xshift=-3ex] {$\sigma$};
  \node [below right] at(uSig.south east) {$1-z_t$};
  \node [below left] at(uSig.south west) {$z_t$};
  \draw[a] (uSig) to (uGate);
  \draw[a] (uSig) to (iGate);
  \node (uDummy) [above=of uSig] {};
  \node (u1) [node distance=1ex,left=of uDummy] {$\W_z\Scalar_t$};
  \node (u2) [node distance=1ex,right=of uDummy] {$\U_z\Hidden_{t-1}$};
  \draw[a] (u1) to (uSig);
  \draw[a] (u2) to (uSig);

  \node (boxdummy) at (Linear.east) {};
  \node[fit=(rGate) (iPsi) (iGate) (boxdummy) (uGate) (uplab)] (box) {};
  \node[below left] at (box.north) {GRU cell};
\end{tikzpicture}

}
  \caption{The gated recurrent unit of \cite{Cho2014}.
    As in the LSTM, the non-linearity $\psi$ is usually $\tanh$.}
  \label{fig:GRU}
\end{figure}
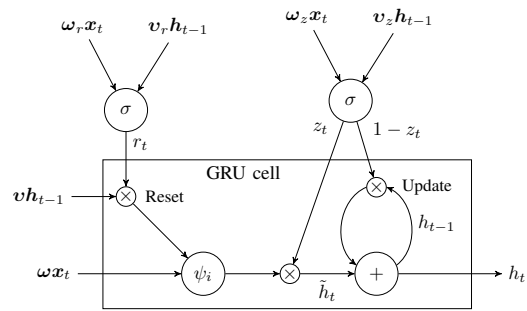

The LSTM's concept of gates has since been used in the gated recurrent unit
(GRU) of Cho et al.~\cite{Cho2014}, and remains important.  In GRU, the input
and forget gates are combined into a single operation, and the output gate is
applied to the recurrent part of the input instead.  It is illustrated in
figure \ref{fig:GRU}.
The GRU has also been modified: In a minimally gated unit (MGU), Zhou et
al.~\cite{Zhou2016a} replace the reset gate with a signal from the update gate;
in the notation here, $r_t$ is replaced by $1-z_t$.  Ravanelli et
al.~\cite{Ravanelli2017, Ravanelli2018} remove the reset gate altogether in
their light GRU (Li-GRU), equivalent to setting $r_t=1$.

Notice that the LSTM and GRU implicitly define three types of recurrence:
\begin{enumerate}
  \item A \emph{unit-wise} recurrence, exemplified by the constant error carousel (CEC, forget loop) of the LSTM or the GRU update loop.
  \item A \emph{layer-wise} recurrence, being the vector loop $\Hidden_{t-1}$  from output to input.
  \item A \emph{gate} recurrence, being the vector loop from output to gate.
\end{enumerate}

Several authors have noted the similarities between HMMs and (recurrent)
networks.  Bourlard and Wellekens \cite{Bourlard1989} show that the two
architectures can be made to compute similar probabilistic values.  Bridle
\cite{Bridle1990c} shows that a suitably designed network can mimic the
\emph{alpha} part of the forward-backward algorithm.  Bridle also points out
similarities between the back-propagation (of derivatives) in the training of
MLPs and the backward pass in HMMs.  With the bidirectional recurrent neural
network (BiRNN), in contrast to \emph{seeking} relationships, Schuster \&
Paliwal \cite{Schuster1997} \emph{imposed} the backward relationship between
HMMs and MLPs by means of a second recurrence relationship running in the
opposite direction.  This was in fact to explicitly allow the network to take
account of ``future'' observations.  The natural substitution of LSTMs for the
same purpose was described by Graves \& Schmidhuber \cite{Graves2005a}
resulting in the bidirectional LSTM (BLSTM or BiLSTM); this type of network
remains the state of the art in several fields.

Putting aside the concept of recurrence, probabilistic interpretations of
feed-forward MLPs are well known.  Although the sigmoid is usually described as
being a smooth (hence differentiable) approximation of a step function, its
probabilistic origin was pointed out by Bridle \cite{Bridle1990a}, and is well
known to physicists via the Boltzmann distribution.  It has also been shown
that the training process yields parameters that make sense in a statistical
sense; this is evident from the work of Richard \& Lippman \cite{Richard1991},
summarising work such as that of \cite{Bourlard1989}, and most thoroughly by
MacKay \cite{MacKay1992a,MacKay1992b,MacKay1992c} in papers that constituted
his PhD thesis, later popularised by Bishop \cite{Bishop1995}.

In the present paper, we build on this latter body of work, recalling that
several MLP concepts have sound Bayesian origins.  We show that this implies a
natural probabilistic recurrence, leading to an architecture similar to the GRU
\cite{Cho2014}.  We go on to show that, because the derivation is
probabilistic, a backward recursion is also evident; this without the explicit
extra backward recurrence of the BiRNN architectures described above.
Experiments on standard speech recognition tasks show that this recurrent
architecture can yield performance near indistinguishable from that of BiRNNs.
Finally, we show that when this implicit bidirectional network is doubled up to
be explicitly bidirectional, it can exceed the performance of BiRNNs.


\section{Background}

\subsection{Bayesian interpretation of MLP units}
\label{sec:BayesBg}


We begin by making explicit a relationship, pointed out by Bridle
\cite{Bridle1990a}, between Bayes's theorem and the sigmoid activation; we show
that the same relationship also applies to ReLU (rectifying linear unit)
activations.

Say we have an observation vector, $\Scalar$, and we want the probability that
it belongs to class $i$, where $i\in\lbrace 1, 2,\dots, C\rbrace$.  The
Bayesian solution is
\begin{equation}
  \label{eq:Bayes}
  \CondPr{\class_i}{\Scalar} =
  \frac
  {\CondLi{\Scalar}{\class_i}\Pr{\class_i}}
  {\sum_{j=1}^C\CondLi{\Scalar}{\class_j}\Pr{\class_j}},
\end{equation}
where $\class_i$ refers to the event that the class takes value $i$, and
$\Scalar$ refers to the event that the observation random variable takes value
$\Scalar$.

If we take the observations to be from multivariate Gaussian distributions
then, in the two class case, $C=2$,
\begin{equation}
  \label{eq:TwoClass}
  \CondPr{\class_1}{\Scalar} =
  \frac{1}{
    1+
    \exp\left(-(\Weight^\T\Scalar + \bias\right))
  },
\end{equation}
where
\begin{align}
  \label{eq:TwoClassWeight}
  \Weight^\T &= (\Mean_1-\Mean_2)^\T\Covariance^{-1} \\
  \label{eq:TwoClassBias}
  \bias &= \nonumber
      \log\Pr{\class_1}
      - \log\Pr{\class_2} \\
             &\qquad -\frac{1}{2}\left(
      \Mean_1^\T\Covariance^{-1}\Mean_1 -
      \Mean_2^\T\Covariance^{-1}\Mean_2
      \right),
\end{align}
and $\Mean_i$ and $\Covariance$ are respectively the mean and covariance of the
constituent Gaussians.  The class priors in this case, $\Pr{\class_i}$, are
taken to be constant and subsumed in the bias term.  This is the commonly used
sigmoid activation.

In the multi-class case, $C\ge 2$,
\begin{equation}
  \CondPr{\class_i}{\Scalar} =
  \frac{
    \exp\left(\Weight_i^\T\Scalar + \bias_i\right)
  }{
    \sum_{j=1}^C
    \exp\left(\Weight_j^\T\Scalar + \bias_j\right)
  },
\end{equation}
where
\begin{align}
  \Weight_i^\T &= \Mean_i^\T\Covariance^{-1} \\
  \label{eq:NClassBias}
  \bias_i &= \log\Pr{\class_i} - \frac{1}{2}\Mean_i^\T\Covariance^{-1}\Mean_i.
\end{align}
This is the softmax activation function introduced in \cite{Bridle1990a}.

A Gaussian assumption is appropriate for MLP inputs.  However, hidden layers
take inputs from previous layers with sigmoid outputs; their values are closer
to beta distributions.  If, instead of a Gaussian, the observations are assumed
to follow independent beta distributions,
\begin{align}
  \Li{\scalar}
  &= \frac{1}{B(\alpha,\beta)}\scalar^{\alpha-1}(1-\scalar)^{\beta-1} \\
  &= \frac{1}{B(\alpha,\beta)}
    e^{(\alpha-1)\log(\scalar)}e^{(\beta-1)\log(1-\scalar)},
\end{align}
where the second line emphasises that the beta is exponential family.  With
$\beta=1$, we then have:
\begin{equation}
  \label{eq:betasigmoid}
  \CondPr{\class_1}{\Scalar} =
  \frac{1}{
    1+
    \exp\left(-(\Weight^\T\log(\Scalar) + \bias)\right)
  },
\end{equation}
with
\begin{align}
  \Alpha_j
  &= (\alpha_{j,1},\dots,\alpha_{j,P})^\T, \\
  \label{eq:Weights}
  \Weight &= \Alpha_1-\Alpha_2 \\
  \bias &=
      \log\Pr{\class_1}
      - \log\Pr{\class_2} \\
      &\quad- \sum_{i=1}^P\left[
        \log B(\alpha_{1,i},1) - \log B(\alpha_{2,i},1)
      \right]
\end{align}
and $P$ is the input dimension.

So, when a sigmoid output is used as the input to a subsequent layer, the value
that makes sense under a beta assumption is its logarithm.  Taking a logarithm
of a sigmoid results in the softplus described by Dugas et al.~\cite{Dugas2001}
albeit for a different reason.  Glorot et al.~\cite{Glorot2011} show that the
ReLU is a linear approximation to the softplus.


\section{General probabilistic recurrence}

In the previous section, we showed that the main activations used in MLPs have
probabilistic explanations.  In this spirit, we derive a recursive activation
from a probabilistic point of view. \rev{At the outset, we expect the formulation to dictate the form of the recursion, removing otherwise ad-hoc aspects of standard techniques.}

\subsection{Conditional independence of observations}

Let us assume that we have a (temporal) sequence of observations
$\Scalar_1,\Scalar_2,\dots,\Scalar_T$.  Equation \ref{eq:Bayes} becomes
(abbreviated for the moment)
\begin{multline}
  \CondPr{\class_i}{\Scalar_T,\Scalar_{T-1},\dots,\Scalar_1}\\
  \propto
  \CondLi{\Scalar_T}{\class_i,\Scalar_{T-1},\dots,\Scalar_1}
  \CondPr{\class_i}{\Scalar_{T-1},\dots,\Scalar_1}.
\end{multline}
If we then assume that all the $\Scalar_t$ are conditionally independent given
$\class_i$, we have
\begin{multline}
  \CondPr{\class_i}{\Scalar_T,\Scalar_{T-1},\dots,\Scalar_1}\\
  \propto
  \CondLi{\Scalar_T}{\class_i}
  \CondPr{\class_i}{\Scalar_{T-1},\dots,\Scalar_1}.
\end{multline}
This is a standard recursion where the posterior at time $t-1$ forms the prior
for time $t$.

\subsection{Application to MLP}

\begin{figure}[htb]
  \centering
  \includegraphics[width=0.9\columnwidth]{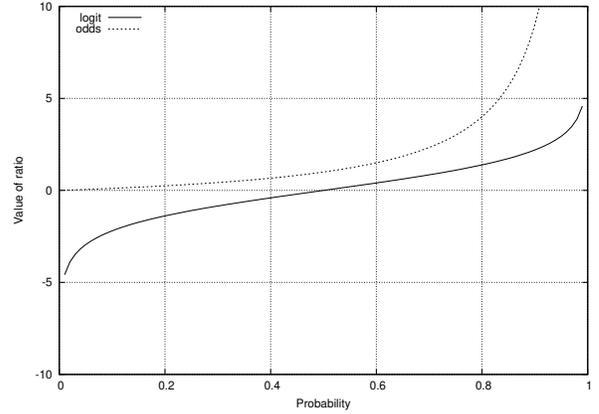}
  \caption{Logit and odds curves.}
  \label{fig:logit}
\end{figure}

More generally, say we have a matrix, $\OBS_T$, the rows of which are
observation vectors $\Obs_1, \Obs_2, \dots, \Obs_T$.  There is a corresponding
matrix, $\HID_T$, the rows of which are vectors
$\Hid_1, \Hid_2, \dots, \Hid_T$.  We assume each element $\hid_{t,i}$ of $\HID$
represents a probability $\CondPr{\feat_i}{\OBS_{t}}$ of the event that feature
$i$ exists in the observation sequence up to time $t$.  Conversely,
$1-\hid_{t,i} = \CondPr{\bar\feat_i}{\OBS_{t}}$.  Notice that, at this stage,
$\feat_i$ is not time dependent; the feature exists (or not) for the whole
sequence, with each observation in the sequence updating
$\CondPr{\feat_i}{\OBS_{t}}$.  Now say that the probabilities
$\CondPr{\feat_i}{\OBS}$ are independent given some parameters, $\Param$.  So
the joint probability is the product
\begin{multline}
  \CondPr{\feat_1,\feat_2,\dots,\feat_F}
  {\Param,\OBS_{t}} = \\
  \CondPr{\feat_1}{\Param,\OBS_{t}}
  \CondPr{\feat_2}{\Param,\OBS_{t}}
  \dots
  \CondPr{\feat_F}{\Param,\OBS_{t}}.
\end{multline}
For a given feature, $\feat_i$,
\begin{align}
  \label{eq:FeatProb}
  \hid_{t,i}
  &=\CondPr{\feat_i}{\Param,\OBS_{t}} \\
  \label{eq:FeatProb2}
  &= \frac{%
    \CondLi{\Obs_t}{\feat_i,\Param,\OBS_{t-1}}
    \CondPr{\feat_i}{\Param,\OBS_{t-1}}
  }{\sum_{\feat_i}
    \CondLi{\Obs_t}{\feat_i,\Param,\OBS_{t-1}}
    \CondPr{\feat_i}{\Param,\OBS_{t-1}}
  } \\ 
  \label{eq:FeatProb3}
  &= \frac{1}{1+
    \dfrac
    {\CondLi{\Obs_t}{\bar\feat_i}}
    {\CondLi{\Obs_t}{\feat_i}}
    \cdot\dfrac{
      \CondPr{\bar\feat_i}{\OBS_{t-1}}
    }{
      \CondPr{\feat_i}{\OBS_{t-1}}
    }
  },
\end{align}
where, in the final line and hereafter, we drop the conditioning on $\Param$
for clarity.  The final expression contains two fractional terms.  The first of
these follows from the conditional independence assumption above, and leads to
the sigmoid of equations \ref{eq:TwoClass} and \ref{eq:betasigmoid}, but
without the priors in the bias terms.  Instead of being static, the priors form
the second fractional term which is a multiplicative feedback
\begin{equation}
  \label{eq:feedback}
  \dfrac{
    \CondPr{\bar\feat_i}{\OBS_{t-1}}
  }{
    \CondPr{\feat_i}{\OBS_{t-1}}
  }
  = \frac{1-\hid_{t-1,i}}{\hid_{t-1,i}}
  = \frac{1}{\odds(\hid_{t-1,i})}
\end{equation}
If this were indeed included as an additive component of the bias in equations
\ref{eq:TwoClass} or \ref{eq:betasigmoid} then the fed back term would be
\begin{align}
  \log\left(\frac{\hid_{t-1,i}}{1-\hid_{t-1,i}}\right)
  &= \logit(\hid_{t-1,i}) \\
  &= \log(\hid_{t-1,i}) - \log(1-\hid_{t-1,i}).
\end{align}
The logit and odds functions are illustrated in figure \ref{fig:logit}.

\section{Probabilistic forget}
\label{sec:forget}

The BRU described above carries the assumption that a feature is present (or
not) in the entire input sequence.  By contrast, we know from the LSTM that it
is necessary to allow an activation to respond differently to different inputs
depending on the context.  In an LSTM this is achieved using gates.  We show
here that gates can be derived probabilistically.

Say that $\Pr{\feat_i}$ is somehow dependent upon another variable indicative
of context.  For instance, if $\feat_i$ is indicative of a characteristic of a
sentence, it is dependent upon the previous words in the sentence, but resets
after a (grammatical) period, when the sentence changes.  Say there is a binary
state variable, $\cvar$, where $\cvar=1$ indicates the context remaining
relevant, and $\cvar=0$ indicates that it is not relevant.  We can assign a
probability, $\cont_t=\CondPr{\cvar_t=1}{\OBS_{t}}$ and the inverse
$(1-\cont_t)=\CondPr{\cvar_t=0}{\OBS_{t}}$, where $\cont_t$ is predicted by the
network.  It is then the prior (in equation \ref{eq:feedback}) that depends on
the context.  $\Feat$ is now dependent upon the time index, $t$.

\rev{Note that the state variable can be defined for one or multiple features.  In the following derivation, we assume only one feature, removing the need for an index.  However, it is common for recurrence to use one variable per feature.}

\subsection{Unit-wise recursion}

We first consider the case where the $\feat_{i}$ are taken to be independent;
it is derived in equations \ref{eq:UnitDerivFirst}--\ref{eq:UnitDerivLast}
below,
\begin{figure*}[!t]
\begin{align}
  \label{eq:UnitDerivFirst}
  \CondPr{\feat_{t,i}}{\OBS_{t-1}}
  &=
    \sum_{\feat_{t-1,i}}
    \sum_{\cvar_{t-1}}
    \CondPr{\feat_{t,i}}{\feat_{t-1,i},\cvar_{t-1},\OBS_{t-1}}
    \CondPr{\feat_{t-1,i}}{\OBS_{t-1}}\CondPr{\cvar_{t-1}}{\OBS_{t-1}} \\
  \begin{split}
    &=
    \CondPr{\feat_{t,i}}{\feat_{t-1,i},\cvar_{t-1}}
    \CondPr{\feat_{t-1,i}}{\OBS_{t-1}}\CondPr{\cvar_{t-1}}{\OBS_{t-1}} \\
    &\quad +
    \CondPr{\feat_{t,i}}{\bar\feat_{t-1,i},\cvar_{t-1}}
    \CondPr{\bar\feat_{t-1,i}}{\OBS_{t-1}}\CondPr{\cvar_{t-1}}{\OBS_{t-1}} \\
    &\quad +
    \CondPr{\feat_{t,i}}{\feat_{t-1,i},\bar\cvar_{t-1}}
    \CondPr{\feat_{t-1,i}}{\OBS_{t-1}}\CondPr{\bar\cvar_{t-1}}{\OBS_{t-1}} \\
    &\quad +
    \CondPr{\feat_{t,i}}{\bar\feat_{t-1,i},\bar\cvar_{t-1}}
    \CondPr{\bar\feat_{t-1,i}}{\OBS_{t-1}}\CondPr{\bar\cvar_{t-1}}{\OBS_{t-1}} \\
  \end{split} \\
  \begin{split}
    &= 1\times \hid_{t-1,i}\cont_{t-1} \\
    &\quad + 0\times (1-\hid_{t-1,i})\cont_{t-1} \\
    &\quad + p_i\hid_{t-1,i}(1-\cont_{t-1}) \\
    &\quad + p_i(1-\hid_{t-1,i})(1-\cont_{t-1}) \\
  \end{split} \\
  &=
    \label{eq:UnitDerivLast}
    (1-\cont_{t-1})p_i + \cont_{t-1}\hid_{t-1,i},
\end{align}
\hrulefill\vspace*{4pt} 
\end{figure*}
where $p_i$ is the unconditional prior probability of feature $i$ being
present.  Notice that the simplifications arise from the interaction of
$\feat_{t,i}$, $\feat_{t-1,i}$ and $\cvar_{t-1}$: context remaining relevant
implies the feature should remain.  So, for instance, the feature changing from
not present to present when context is relevant has zero probability.

In a Kalman filter sense, $\CondPr{\feat_{t,i}}{\OBS_{t-1}}$ is the predictor.
The result is an intuitive linear combination of the previous output with a
prior.  In this paper, although we deal with a discrete state variable, we use
the Kalman filter analogy because it is easier to follow.  Nevertheless, a
correspondence with \emph{alpha}, \emph{beta} and \emph{gamma} probabilities
will be evident to readers familiar with Markov models.

There is a question of initialisation.  The first output corresponding to $t=1$
should use the value $\hid_{0,i}=p_i$; thereafter, the value from the feedback
loop can be taken.

\begin{figure}[htb]
  \centering
  \resizebox{0.6\columnwidth}{!}{\begin{tikzpicture}[every fit/.style={rectangle,draw,inner sep=1.5ex}]

  \node (Sigmoid) [c] {$\sigma$};
  \node (iScalar) [node distance=10ex, left=of Sigmoid] {$\W\Scalar_t$};
  \node (iHidden) [below=of iScalar] {$\U\Hidden_{t-1}$};
  \draw[a] (iScalar) to (Sigmoid);
  \draw[a] (iHidden) to (Sigmoid);
  
  \node (uLog) [c,above=of Sigmoid] {$f(\cdot)$};
  \draw[a] (Sigmoid) to [bend right=60] node (d1) {} (uLog.east);
  \draw[a] (uLog.west) to [bend right=60] node (d2) {} (Sigmoid);

  \node (output) [node distance=10ex, right=of Sigmoid] {$h_t$};
  \draw[a] (Sigmoid) to (output);

  \node (fSig) [node distance=10ex, c,above=of uLog] {$\sigma$};
  \node [below right] at (fSig.south) {$z_{t-1}$};
  \draw[a] (fSig) to (uLog);
  \node (dummy) [above=of fSig] {};
  \node (f1) [node distance=1ex,left=of dummy] {$\W_f\Scalar_{t-1}$};
  \node (f2) [node distance=1ex,right=of dummy] {$\U_f\Hidden_{t-2}$};
  \draw[a] (f1) to (fSig);
  \draw[a] (f2) to (fSig);

  \node (prior) [node distance=3ex, above left=of uLog] {$p$};
  \draw [a] (prior) to (uLog);

  \node[fit=(uLog) (Sigmoid) (prior) (d1) (d2)] (box) {};
  \node[below right] at (box.north east) {BRU cell};
\end{tikzpicture}

}
  \caption{A Bayesian recurrent unit incorporating a probabilistic forget
    gate.}
  \label{fig:BRAU}
\end{figure}
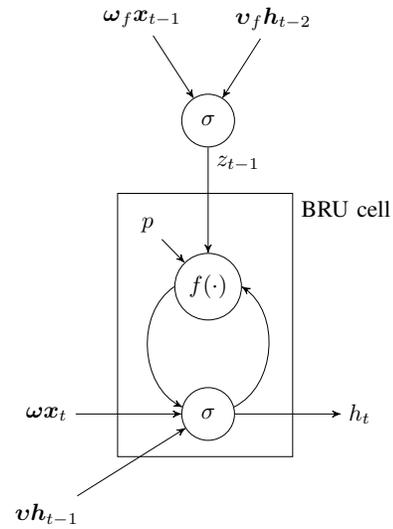

\begin{tabular}{lll}
  At time $t=1$, & $\cont_{t-1}=0$, & $\hid_{t-1,i}=p_i$ \\
  At time $t=2$, & $\cont_{t-1}=f_\cont(\OBS_{t-1})$,
  & $\hid_{t-1,i}=f_\hid(\OBS_{t-1})$ \\
\end{tabular}

where $f_\cdot(\cdot)$ is taken to mean ``some function of''.  \rev{If
  $\hid_{t-2,i}$ is required, the same value as $\hid_{t-1,i}$ can be used.}
In turn, the fed back value (equation \ref{eq:feedback}) is actually
\begin{equation}
  \frac{1-\CondPr{\feat_{t,i}}{\OBS_{t-1}}}
  {\CondPr{\feat_{t,i}}{\OBS_{t-1}}}
  =
  \frac{1}{\odds\left([1-\cont_{t-1}]p_i + \cont_{t-1}\hid_{t-1,i}\right)},
\end{equation}
with the logarithm of the reciprocal being the additive term inside the
exponential.  This is illustrated in figure \ref{fig:BRAU} where,
\begin{equation}
  \label{eq:UnitLogit}
  f(\cdot) = \logit\left([1-\cont_{t-1}]p_i + \cont_{t-1}\hid_{t-1,i}\right).
\end{equation}
In figure \ref{fig:BRAU}, note that the unit-wise recurrence is probabilistic,
but an \emph{ad-hoc} layer-wise and gate recurrence are also retained for
comparison with a GRU.  The $\Hidden_{t-2}$ term in this and later cases arises
to maintain a consistent definition of $\cont_t$ across the LSTM, GRU and
equation \ref{eq:summary1}; we note that, in practice, the extra delay makes no
difference in performance.


\subsection{Discussion}

The unit-wise recursion above was an attempt to formalise the ``constant error
carousel'' (CEC) --- the central recurrence --- of the LSTM.  Whilst the result
is self consistent, in practice we find two difficulties:
\begin{enumerate}
\item The logit function of equation \ref{eq:UnitLogit} causes instability in
  the training process.  This is because it can tend to $\pm\infty$.
\item The formulation does not explain the layer-wise recursion around the
  whole layer of units.
\end{enumerate}
In the following, we address both of these difficulties using approximations.
We find that the resulting layer-wise recursion is both stable and more
complete.

\subsection{Layer-wise recursion}
\label{sec:layer-forward}
In contrast to the unit-wise recursion above, here we take the elements of
$\Feat$ to be dependent, meaning the summation is over the whole vector.  The
main derivation is equations \ref{eq:LayerDerivFirst}--\ref{eq:LayerDerivLast}
below,
\begin{figure*}[!t]
\begin{align}
  \label{eq:LayerDerivFirst}
  \CondPr{\feat_{t,i}}{\OBS_{t-1}}
  &=
    \sum_{\Feat_{t-1}}
    \sum_{\cvar_{t-1}}
    \CondPr{\feat_{t,i}}{\Feat_{t-1},\cvar_{t-1},\OBS_{t-1}}
    \CondPr{\Feat_{t-1}}{\OBS_{t-1}}\CondPr{\cvar_{t-1}}{\OBS_{t-1}} \\
  \begin{split}
    &=
    \CondPr{\cvar_{t-1}}{\OBS_{t-1}}\sum_{\Feat_{t-1}}
    \CondPr{\feat_{t,i}}{\Feat_{t-1},\cvar_{t-1}}
    \CondPr{\Feat_{t-1}}{\OBS_{t-1}} \\
    &\quad +
    \CondPr{\bar\cvar_{t-1}}{\OBS_{t-1}}\sum_{\Feat_{t-1}}
    \CondPr{\feat_{t,i}}{\Feat_{t-1},\bar\cvar_{t-1}}
    \CondPr{\Feat_{t-1}}{\OBS_{t-1}}.
  \end{split}\\
  \label{eq:LayerDerivLast}
  \begin{split}
    &= \cont_{t-1}\sum_{\Feat_{t-1}}
    \CondPr{\feat_{t,i}}{\Feat_{t-1},\cvar_{t-1}}
    \prod_{i}\CondPr{\feat_{t-1,i}}{\OBS_{t-1}} \\
    &\quad +
    (1-\cont_{t-1})\sum_{\Feat_{t-1}}
    \CondPr{\feat_{t,i}}{\Feat_{t-1},\bar\cvar_{t-1}}
    \prod_{i}\CondPr{\feat_{t-1,i}}{\OBS_{t-1}}.
  \end{split}
\end{align}
\hrulefill\vspace*{4pt} 
\end{figure*}
The calculation can be rendered tractable if we model $\CondPr{\feat_{t,i}}{\Feat_{t-1},\cvar_{t-1}}$ as $\Weight_i^\T\Feat_{t-1}$, where $\Weight_i$ is a trainable vector and each element $\weight_{j,i}$ models the weight that $\feat_{t-1,j}$ has on $\feat_{t,i}$. The occurrence of $\feat_{t,i}$ is considered to be the weighted average of the occurrences of $\Feat_{t-1}$. This is an extension of unit-wise recursion where the occurrence of $\feat_{t,i}$ only depends on $\feat_{t-1,i}$ and $\Weight_i$ is a one-hot vector with $\weight_{i,i}=1$. Therefore, we have
\begin{equation}
  \sum_{\Feat_{t-1}}\CondPr{\feat_{t,i}}{\Feat_{t-1},\cvar_{t-1}}\prod_{i}\CondPr{\feat_{t-1,i}}{\OBS_{t-1}} = \Weight_i^\T\Hid_t+c
\end{equation}
where, $\Weight_i$ denotes the $i^{th}$ column of $\Weight$ and $c=\sum_{j\in\{j|\weight_{j,i}<0\}}\weight_{j,i}$. To understand the above equation, consider $N$ independent lotteries, where $N$ is the total number of nodes in a layer. The winning rate of the $i^{th}$ lottery is $\hid_i$, the corresponding prize is $\weight_i$. Now we buy each of the lottery once. The left side of the above equation actually calculate the expectation of the total prizes we can win from the lotteries by listing all the possibilities. On the other hand, each lottery is independent. Therefore, the expectation prize for $i^{th}$ lottery is $\weight_i\hid_i$. The expectation of the total prizes we can get is then $\Weight_i^\T\Hid$.

\begin{figure}[htb]
  \centering
  \resizebox{0.7\columnwidth}{!}{\begin{tikzpicture}[every fit/.style={rectangle,draw,inner sep=1.5ex}]

  \foreach \s [count=\c] in {0.8,0.7,0.6,0.5,0.4,0.3,0.2,0.1}
    \node (Sigmoid-\c) [c,fill=white] at +(\s,\s) {$\sigma$};

  \node (Sigmoid) [c,fill=white] {$\sigma$};
  \node (iScalar) [node distance=10ex, left=of Sigmoid] {$\W\Scalar_t$};
  \draw[a] (iScalar) to (Sigmoid);

  \node (uLog) [c,above=of Sigmoid] {$f(\cdot)$};
  \foreach \s [count=\c] in {0.8,0.7,0.6,0.5,0.4,0.3,0.2,0.1}
    \node (uLog-\c) [c,fill=white] at ($(uLog) +(\s,\s)$) {$f(\cdot)$};

  \node [c,fill=white,above=of Sigmoid] {$f(\cdot)$};
  \draw[a] (Sigmoid) to [bend right=80] (uLog.east);
  \foreach \c in {3,5,7}
    \draw[a] (Sigmoid-\c) to [bend right=80] (uLog.east);
  \draw[a] (Sigmoid-1) to [bend right=80] node (d1) [right] {$\U\Hidden_{t-1}$} (uLog.east);
  \draw[a] (uLog.west) to [bend right=60] node (d2) {} (Sigmoid);

  \node (output) [node distance=17ex, right=of Sigmoid] {$\Hidden_t$};
  \draw[a] (Sigmoid) to (output);

  \node (fSig) [node distance=10ex, c,above=of uLog] {$\sigma$};
  \node [below right] at (fSig.south) {$z_{t-1}$};
  \draw[a] (fSig) to (uLog);
  \node (dummy) [above=of fSig] {};
  \node (f1) [node distance=1ex,left=of dummy] {$\W_f\Scalar_{t-1}$};
  \node (f2) [node distance=1ex,right=of dummy] {$\U_f\Hidden_{t-2}$};
  \draw[a] (f1) to (fSig);
  \draw[a] (f2) to (fSig);

  \node (prior) [node distance=3ex, above left=of uLog] {$p$};
  \draw [a] (prior) to (uLog);

  \node[fit=(uLog) (Sigmoid) (uLog-1) (prior) (d1) (d2)] (box) {};
  \node[above left] at (box.south east) {LBRU cell};
\end{tikzpicture}

}
  \caption{The layer-wise recursion with a forget gate.}
  \label{fig:FBRU}
\end{figure}
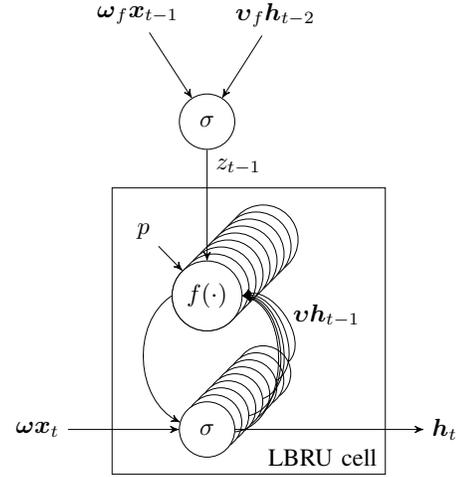

In this sense, the recursion is parameterised by matrix $\Weight$.  Given the fact that $\Weight_i^\T\Feat_{t-1}$ represents probabilities and the expectation of probabilities should be positive, it is sensible to constrain the $L_1$ norm of each column in $\Weight$ to $1$ and add the bias term $c$. Thus, equation \ref{eq:LayerDerivLast} can be written as
\begin{align}
  \nonumber
  &\CondPr{\feat_{t,i}}{\OBS_{t-1}}
 = \cont_{t-1}(\Weight_i^\T\Hid_{t-1}+c-p_i) + p_i
\end{align}
This is illustrated in \ref{fig:FBRU}, where,
\begin{equation}
  f(\cdot) = \logit\left(\cont_{t-1}(\Weight_i^\T\Hid_{t-1}+c-p_i)+p_i\right).
\end{equation}
Note that in figure \ref{fig:FBRU}, the unit-wise and layer-wise recurrence are combined into a single probabilisitic recurrence.  However, the ad-hoc gate recurrence is retained.
With reference to figure \ref{fig:logit}, the function
$\log\left(\frac{\hid}{1-\hid}\right)$ appears linear except for narrow regions close to $0$ and $1$.  Since we are not aware of the distribution of $\hid$, we further approximate $\log\left(\frac{\hid}{1-\hid}\right) \approx \alpha\hid+\beta$, yielding
\begin{equation}
  \log\left(
    \frac{\CondPr{\feat_{t,i}}{\OBS_{t-1}}}
    {1-\CondPr{\feat_{t,i}}{\OBS_{t-1}}}
  \right)
  \approx
  \cont_{t-1}(\Weight_i^\T\Hid_{t-1} +c- p_i) + p_i+\beta,
\end{equation}
where $\alpha$ is absorbed by $\Weight_i^\T$ and $p_i$. The range of $\alpha$ is $[4,+\infty)$. Therefore, we do not normalise $\Weight_i$ in the forward pass.

Substituting back into equation \ref{eq:FeatProb3}, that equation can be rewritten as:
\begin{align}
  \Hid_{t}
  &=\sigma(\Weight_{ih} \Obs_t + \mathbf{b}_{ih} + \mathbf{\cont}_{t-1}\odot(\Weight_{hh} \Hid_{t-1}+ \mathbf{b}_{hh}))
\end{align}
which is quite similar to the function of the reset gate in a GRU:
\begin{equation}
  \label{eq:gru}
  \mathbf{n}_t =
  \tanh(\Weight_{in} \Obs_t + \mathbf{b}_{in} + \mathbf{r}_t\odot (\Weight_{hn} \Hid_{t-1}+ \mathbf{b}_{hn}))
\end{equation}
Besides the activation function, another main difference is that the forget
gate
$\cont_{t-1}$ is computed in the previous time step. If $\cont_{t-1}$ degrades to a constant $1$, we get the formulation of a basic recurrent layer that is used in practice.

\section{Backward recursion}

The recursions described thus far only yield accurate probabilities at time
$t=T$.  The earlier ones ($1<t<T$) depend upon future observations.  This is
normally corrected via the backward passes of either the Kalman smoother or
forward-backward algorithm.  In this section, we derive backward recursions for
the recurrent units derived above.  In fact, the ability to do this is one of
the most compelling reasons to derive probabilistic recurrence.

\subsection{Unit-wise recursion}

Although the unit-wise recurrence (without approximations) is unstable, it
turns out to be beneficial (see section \ref{sec:experiments}) to derive the
backward pass.  It can be done without adding extra parameters, making it
directly comparable to the GRU.

Following the method for the Kalman smoother, we
first integrate over the state at time $t$ and the context variable,
\begin{align}
  \nonumber
  &\CondPr{\feat_{t-1,i}}{\OBS_{t}} \\
  &\quad =
    \sum_{\feat_{t,i},\cvar_{t-1}}
    \CondPr{\feat_{t-1,i}}{\feat_{t,i},\cvar_{t-1},\OBS_{t}}
    \CondPr{\feat_{t,i},\cvar_{t-1}}{\OBS_{t}} \\
  \begin{split}
    &\quad =
    \CondPr{\feat_{t-1,i}}{\feat_{t,i},\cvar_{t-1},\OBS_{t}}
    \hid_{t,i}\cont_{t-1}\\
    &\qquad +
    \CondPr{\feat_{t-1,i}}{\bar\feat_{t,i},\cvar_{t-1}\OBS_{t}}
    (1-\hid_{t,i})\cont_{t-1}\\
    &\qquad +
    \CondPr{\feat_{t-1,i}}{\feat_{t,i},\bar\cvar_{t-1},\OBS_{t}}
    \hid_{t,i}(1-\cont_{t-1})\\
    &\qquad +
    \CondPr{\feat_{t-1,i}}{\bar\feat_{t,i},\bar\cvar_{t-1}\OBS_{t}}
    (1-\hid_{t,i})(1-\cont_{t-1}).
  \end{split}
\end{align}
Note that, given $\feat_{t,i}$, $\Pr{\feat_{t-1,i}}$ is conditionally
independent of any data after time $t-1$.  Equations
\ref{eq:UnitBackwardFirst}--\ref{eq:UnitBackwardLast} show how to use Bayes's
theorem to expand the remaining terms.
\begin{figure*}
\begin{align}
  \label{eq:UnitBackwardFirst}
  \CondPr{\feat_{t-1,i}}{\feat_{t,i},\cvar_{t-1},\OBS_{t}}
  &=
    \frac{
    \CondPr{\feat_{t,i}}{\feat_{t-1,i},\cvar_{t-1}}
    \CondPr{\feat_{t-1,i}}{\OBS_{t-1}}
    }{
    \sum_{\feat_{t-1,i}}
    \CondPr{\feat_{t,i}}{\feat_{t-1,i},\cvar_{t-1}}
    \CondPr{\feat_{t-1,i}}{\OBS_{t-1}}
    } \\
  &= \frac{1\times\hid_{t-1,i}}{1\times\hid_{t-1,i} + 0\times(1-\hid_{t-1,i})}
    = 1. \\
  \CondPr{\feat_{t-1,i}}{\bar\feat_{t,i},\cvar_{t-1},\OBS_{t}}
  &=
    \frac{
    \CondPr{\bar\feat_{t,i}}{\feat_{t-1,i},\cvar_{t-1}}
    \CondPr{\feat_{t-1,i}}{\OBS_{t-1}}
    }{
    \sum_{\feat_{t-1,i}}
    \CondPr{\bar\feat_{t,i}}{\feat_{t-1,i},\cvar_{t-1}}
    \CondPr{\feat_{t-1,i}}{\OBS_{t-1}}
    } \\
  &= \frac{0\times\hid_{t-1,i}}{0\times\hid_{t-1,i} + 1\times(1-\hid_{t-1,i})}
    = 0. \\
  \CondPr{\feat_{t-1,i}}{\feat_{t,i},\bar\cvar_{t-1},\OBS_{t}}
  &=
    \frac{
    \CondPr{\feat_{t,i}}{\feat_{t-1,i},\bar\cvar_{t-1}}
    \CondPr{\feat_{t-1,i}}{\OBS_{t-1}}
    }{
    \sum_{\feat_{t-1,i}}
    \CondPr{\feat_{t,i}}{\feat_{t-1,i},\bar\cvar_{t-1}}
    \CondPr{\feat_{t-1,i}}{\OBS_{t-1}}
    } \\
  &= \frac{p_i\hid_{t-1,i}}{p_i\hid_{t-1,i} + p_i(1-\hid_{t-1,i})}
    = \hid_{t-1,i}. \\
  \CondPr{\feat_{t-1,i}}{\bar\feat_{t,i},\bar\cvar_{t-1},\OBS_{t}}
  &=
    \frac{
    \CondPr{\bar\feat_{t,i}}{\feat_{t-1,i},\bar\cvar_{t-1}}
    \CondPr{\feat_{t-1,i}}{\OBS_{t-1}}
    }{
    \sum_{\feat_{t-1,i}}
    \CondPr{\bar\feat_{t,i}}{\feat_{t-1,i},\bar\cvar_{t-1}}
    \CondPr{\feat_{t-1,i}}{\OBS_{t-1}}
    } \\
  \label{eq:UnitBackwardLast}
  &= \frac{(1-p_i)\hid_{t-1,i}}{(1-p_i)\hid_{t-1,i} + (1-p_i)(1-\hid_{t-1,i})}
    = \hid_{t-1,i}.
\end{align}
\hrulefill\vspace*{4pt} 
\end{figure*}
Putting the above together, we initialise
\begin{equation}
  \hid_{T,i}' = \hid_{T,i}
\end{equation}
then recurse
\rev{
\begin{align}
  \hid'_{t-1}
  &= \CondPr{\feat_{t-1,i}}{\OBS_{t}} \\
  &= \nonumber
    \hid_{t,i}'\cont_{t-1} +
    \hid_{t-1,i}\hid_{t,i}'(1-\cont_{t-1})\\
  &\quad +
    \hid_{t-1,i}(1-\hid_{t,i}')(1-\cont_{t-1}) \\
  &= (1-\cont_{t-1})\hid_{t-1,i} + \cont_{t-1}\hid_{t,i}'.
\end{align}}

\subsection{Layer-wise recursion}

Now we consider the case that $\feat_{t-1,i}$ is dependent on the whole vector $\Feat_t$.
\begin{align}
  \nonumber
  &\CondPr{\feat_{t-1,i}}{\OBS_{t}} \\
  &\quad =
    \sum_{\Feat_{t},\cvar_{t-1}}
    \CondPr{\feat_{t-1,i}}{\Feat_{t},\cvar_{t-1},\OBS_{t}}
    \CondPr{\Feat_{t},\cvar_{t-1}}{\OBS_{t}} \\
  \begin{split}
    &\quad =
   \cont_{t-1} \sum_{\Feat_{t}}\CondPr{\feat_{t-1,i}}{\Feat_{t},\cvar_{t-1},\OBS_{t}}
    \CondPr{\Feat_{t}}{\OBS_{t}}\\
    &\qquad +
    (1-\cont_{t-1})\sum_{\Feat_{t}}\CondPr{\feat_{t-1,i}}{\Feat_{t},\bar\cvar_{t-1},\OBS_{t}}
     \CondPr{\Feat_{t}}{\OBS_{t}}\\
  \end{split}
\end{align}
Equations \ref{eq:LayerBackwardFirst}--\ref{eq:LayerBackwardLast} show how to
use use Bayes's theorem to expand the remaining terms,
\begin{figure*}[!t]
\begin{align}
  \label{eq:LayerBackwardFirst}
  \CondPr{\feat_{t-1,i}}{\Feat_{t},\cvar_{t-1},\OBS_{t}}
  &= \frac{\CondPr{\Feat_{t}}{\feat_{t-1,i},\cvar_{t-1}}\CondPr{\feat_{t-1,i}}{\OBS_{t-1}}}{\sum_{\Feat_{t-1}}\CondPr{\Feat_t}{\Feat_{t-1},\cvar_{t-1}}\CondPr{\Feat_{t-1}}{\OBS_{t-1}}} \\
  &= \frac{\sum_{\Feat_{t-1,\bar i}}\CondPr{\Feat_{t}}{\feat_{t-1,i},\Feat_{t-1, \bar i},\cvar_{t-1}}\CondPr{\Feat_{t-1,\bar i}}{\OBS_{t-1}}\CondPr{\feat_{t-1,i}}{\OBS_{t-1}}}{\sum_{\Feat_{t-1}}\CondPr{\Feat_t}{\Feat_{t-1},\cvar_{t-1}}\CondPr{\Feat_{t-1}}{\OBS_{t-1}}}
\end{align}
\begin{align}
  \CondPr{\feat_{t-1,i}}{\Feat_{t},\bar \cvar_{t-1},\OBS_{t}}
  &= \frac{\sum_{\Feat_{t-1,\bar i}}\CondPr{\Feat_{t}}{\feat_{t-1,i},\Feat_{t-1,\bar i},\bar \cvar_{t-1}}\CondPr{\Feat_{t-1,\bar i}}{\OBS_{t-1}}\CondPr{\feat_{t-1,i}}{\OBS_{t-1}}}{\sum_{\Feat_{t-1}}\CondPr{\Feat_t}{\Feat_{t-1},\bar \cvar_{t-1}}\CondPr{\Feat_{t-1}}{\OBS_{t-1}}} \\
  &= \frac{\prod_k p_k}{\prod_k p_k(1+(1-\hid_{t-1,i})/\hid_{t-1,i})} \\
  \label{eq:LayerBackwardLast}
  &= \hid_{t-1,i}
\end{align}
\hrulefill\vspace*{4pt} 
\end{figure*}
where $\Feat_{t-1, \bar i}$ denotes the features of all the units in the layer except the $i^{th}$ unit and $p_k$ is the prior probability of unit $k$. The first term $\CondPr{\feat_{t-1,i}}{\Feat_{t},\cvar_{t-1},\OBS_{t}}$ seems intractable, although it allows us to re-use the weights learnt from the forward pass to smooth the output via backward recursion. Now suppose there is another binary state variable, $\xi_t$, where $\xi_t=1$ indicates the future context remaining relevant, meaning that $\feat_t$ is dependent on $\feat_{t+1}$ and $\xi=0$ indicates that the future context is irrelevant.  We can assign a new probability, $\back_t=\CondPr{\xi_t=1}{\OBS_{t}}$ and the inverse $(1-\back_t)=\CondPr{\xi_t=0}{\OBS_{t}}$. We assume $\xi_t$ is independent of future observations $\OBS_{t+1}^{T}$. Thus, we can write:
%
%
\begin{align}
  \nonumber
  &\CondPr{\feat_{t-1,i}}{\OBS_{T}} \\
  &\quad =
    \sum_{\Feat_{t},\xi_{t}}
    \CondPr{\feat_{t-1,i}}{\Feat_{t},\xi_{t-1},\OBS_{T}}
    \CondPr{\Feat_{t},\xi_{t-1}}{\OBS_{T}} \\
  \begin{split}
    &\quad =
    \back_{t-1}\sum_{\Feat_{t}}\CondPr{\feat_{t-1,i}}{\Feat_{t},\xi_{t-1}}
    \prod_{k}\CondPr{\feat_{t,k}}{\OBS_T}\\
    &\qquad +
    (1-\back_{t-1})\sum_{\Feat_{t}}\CondPr{\feat_{t-1,i}}{\bar\xi_{t-1},\OBS_{t-1}}
    \CondPr{\Feat_{t}}{\OBS_T}
  \end{split} \\
    &\quad =  \back_{t-1}\sum_{\Feat_{t}}\CondPr{\feat_{t-1,i}}{\Feat_{t},\xi_{t-1}}\prod_{k}\CondPr{\feat_{t,k}}{\OBS_T} \\
    &\qquad + \hid_{t-1,i}(1-\back_{t-1}).
\end{align}
Similarly, we model $\CondPr{\feat_{t-1,i}}{\Feat_{t},\xi_{t-1}}$ as $\Weight_i^\T\Feat_{t}$, the product of a trainable vector $\Weight_i$ and $\Feat_t$, and denote $\hid'_{t,i} = \CondPr{\feat_{t,i}}{\OBS_{T}}$ and put the above together, we initialise
\begin{equation}
  \hid_{T,i}' = \hid_{T,i}
\end{equation}
then recurse
\begin{align}
 \CondPr{\feat_{t-1,i}}{\OBS_{T}}= \hid'_{t-1,i}
  &= \back_t(\Weight_i^\T\Hid_t'+c) + \hid_{t-1,i}(1-\back_{t}),
\end{align}
where $c=\sum_{j\in\{j|\weight_{j,i}<0\}}\weight_{j,i}$. It is sensible to apply the same constraints discussed in Section \ref{sec:layer-forward} to the backward recurrent matrix and add the bias term.

The layer-wise backward pass hence requires extra parameters.  In this sense it
is not directly comparable to a similar GRU.  Nevertheless, the parameter count
is smaller than for a bidirectional GRU.  The repercussions of this are
examined in section \ref{sec:experiments}.

\section{Probabilistic input}

In examining the probabilistic forget derivations above, whilst we set out to
formalise the CEC of the LSTM, the result is closer to the reset gate of a GRU.
In this section, we show that the update gate of a GRU can also be derived
rather simply.

\subsection{Recursion}

\begin{figure}[htb]
  \centering
  \resizebox{0.8\columnwidth}{!}{\begin{tikzpicture}[every fit/.style={rectangle,draw,inner sep=1.5ex}]

  \foreach \s [count=\c] in {0.8,0.7,0.6,0.5,0.4,0.3,0.2,0.1}
    \node (Sigmoid-\c) [c,fill=white] at +(\s,\s) {$\sigma$};
  
  \node (Sigmoid) [c,fill=white] {$\sigma$};
  \node (iScalar) [node distance=10ex, left=of Sigmoid] {$\W\Scalar_t$};
  \draw[a] (iScalar) to (Sigmoid);

  \node (uLog) [c,above=of Sigmoid] {$f(\cdot)$};
  \foreach \s [count=\c] in {0.8,0.7,0.6,0.5,0.4,0.3,0.2,0.1}
    \node (uLog-\c) [c,fill=white] at ($(uLog) +(\s,\s)$) {$f(\cdot)$};
  \node [c,fill=white,above=of Sigmoid] {$f(\cdot)$};
  \draw[a] (uLog.west) to [bend right=60] node (d2) {} (Sigmoid);
  
  \node (fSig) [node distance=10ex, c,above=of uLog] {$\sigma$};
  \node [below right] at (fSig.south) {$z_{t-1}$};
  \draw[a] (fSig) to (uLog);
  \node (dummy) [above=of fSig] {};
  \node (f1) [node distance=1ex,left=of dummy] {$\W_f\Scalar_{t-1}$};
  \node (f2) [node distance=1ex,right=of dummy] {$\U_f\Hidden_{t-2}$};
  \draw[a] (f1) to (fSig);
  \draw[a] (f2) to (fSig);

  \node (prior) [node distance=3ex, above left=of uLog] {$p$};
  \draw [a] (prior) to (uLog);

  \node (iGate) [op,right=of Sigmoid] {$\times$};

  \node (Linear) [c,right=of iGate] {$+$};
  \foreach \s [count=\c] in {0.8,0.7,0.6,0.5,0.4,0.3,0.2,0.1}
    \node (Linear-\c) [c,fill=white] at ($(Linear) +(\s,\s)$) {$+$};
  \node [c,fill=white,right=of iGate] {$+$};
  \draw[a] (Sigmoid) to (iGate);
  \draw[a] (iGate) to (Linear);

  \draw[a] (Linear) to [bend right=40] (uLog.east);
  \foreach \c in {3,5,7}
    \draw[a] (Linear-\c) to [bend right=40] (uLog.east);
  \draw[a] (Linear-1) to [bend right=40] node (d1) [above] {$\U\Hidden_{t-1}$} (uLog.east);

  \node (uGate) [op,below=of Linear] {$\times$};
  \draw[a] (Linear) to [bend left=60] node (d3) [right] {$h_{t-1}$} (uGate.east);
  \draw[a] (uGate.west) to [bend left=60] (Linear);

  \node (uSig) [c,below=of uGate,xshift=-3ex] {$\sigma$};
  \node [above right] at(uSig.north east) {$1-r_t$};
  \node [above left] at(uSig.north west) {$r_t$};
  \draw[a] (uSig) to (uGate);
  \draw[a] (uSig) to (iGate);
  \node (uDummy) [below=of uSig] {};
  \node (u1) [node distance=1ex,left=of uDummy] {$\W_i\Scalar_t$};
  \node (u2) [node distance=1ex,right=of uDummy] {$\U_i\Hidden_{t-1}$};
  \draw[a] (u1) to (uSig);
  \draw[a] (u2) to (uSig);

  \node (output) [node distance=12ex,right=of Linear] {$\Hidden_t$};
  \draw[a] (Linear) to (output);

  \node[fit=(uLog) (Linear) (uGate) (uLog-1) (prior) (d1) (d2) (d3)] (box) {};
  \node[below left] at (box.north east) {LBRU cell};
\end{tikzpicture}

}
  \caption{The layer-wise recursion with a forget gate and an input gate.}
  \label{fig:IFBRU}
\end{figure}
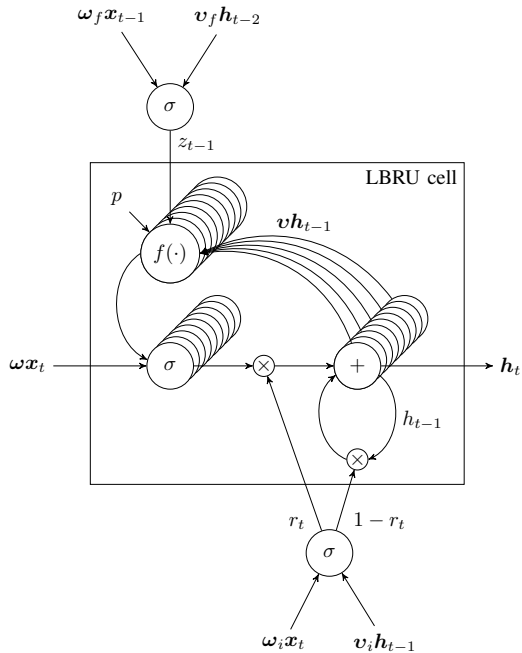

In the same spirit as the previous section, say there is a binary state
variable, $\rvar$, where $\rvar=1$ indicates the current input is relevant, and
$\rvar=0$ indicates that it is not relevant. We can assign a probability,
$\rlvn_t=\CondPr{\rvar_t=1}{\OBS_{t}}$ and the inverse
$(1-\rlvn_t)=\CondPr{\rvar_t=0}{\OBS_{t}}$. We assume if the current input is
irrelevant, then $\feat_t$ is completely dependent on $\feat_{t-1}$.  For a
given feature, $\feat_i$, the derivation is shown in equations
\ref{eq:ProbInputFirst}--\ref{eq:ProbInputLast} below.
\begin{figure*}
\rev{\begin{align}
  \label{eq:ProbInputFirst}
  \hid_{t,i}
  &= \CondPr{\feat_{t,i}}{\OBS_{t}} \\
  \label{eq:input}
  &= \sum_{\rvar_{t,i}} \CondPr{\feat_{t,i}}{\OBS_{t},\rvar_{t,i}}\CondPr{\rvar_{t,i}}{\OBS_t} \\
  &= \CondPr{\feat_{t,i}}{\OBS_{t},\rvar_{t,i}}\CondPr{\rvar_{t,i}}{\OBS_t}+	\sum_{\feat_{t-1,i}}\CondPr{\feat_{t,i}}{\OBS_{t},\feat_{t-1,i},\bar{\rvar}_{t,i}}\CondPr{\bar{\rvar}_{t,i}}{\OBS_t}\CondPr{\feat_{t-1,i}}{\OBS_t} \\
    &\approx \CondPr{\feat_{t,i}}{\OBS_{t}}\CondPr{\rvar_{t,i}}{\OBS_t} +\sum_{\feat_{t-1,i}}\CondPr{\feat_{t,i}}{\feat_{t-1,i},\bar{\rvar}_{t,i}}\CondPr{\bar{\rvar}_{t,i}}{\OBS_t}\CondPr{\feat_{t-1,i}}{\OBS_{t-1}}\\
  \label{eq:ProbInputLast}
  &= \rlvn_{t,i}\CondPr{\feat_{t,i}}{\OBS_{t}} + (1-\rlvn_{t,i})\hid_{t-1,i}
\end{align}}
\hrulefill\vspace*{4pt} 
\end{figure*}
The first term follows the same derivations in previous sections. This is illustrated in Fig. \ref{fig:IFBRU}, where, as before, the unit-wise and layer-wise recursions are merged, and the gate recursion remains ad-hoc; this provides for a fair comparison with GRU in section \ref{sec:experiments}.

This correlates to the update function in a GRU:
\begin{equation}
  \Hid_t = (1 - \mathbf{z}_t) \odot\mathbf{n}_t + \mathbf{z}_t \Hid_{(t-1)},
\end{equation}
where $n_t$ is defined as equation \ref{eq:gru} and $z_t$ is the update gate
computed as
\begin{equation}
  \mathbf{z}_t = \sigma(\Weight_{iz} \Obs_t + \mathbf{b}_{iz} + \Weight_{hz} \Hid_{(t-1)} + \mathbf{b}_{hz})
\end{equation}

It may be argued that the
input gate and the forget gate have simliar functionality. Indeed, if we only
keep the forget gate and let $\cont_t=\CondPr{\rvar_t=0}{\OBS_{t}}$, this leads
to the MGU \cite{Zhou2016a}; If we keep the forget gate always equal to $1$, it
leads to the Li-GRU \cite{Ravanelli2018}.

We do not derive a backward recursion for the input gate.  Rather, the
resulting resemblance to the GRU provides us with a candidate architecture to
compare experimentally; this is reported in section \ref{sec:experiments}.

\rev{%
\subsection{Summary}

The forward pass of this final BRU can be summarised as:
\begin{align}
  \label{eq:summary1}
  \mathbf{\cont}_{t}&=\sigma(\Weight_{iz}\Obs_t  + \Weight_{hz}\Hid_{t-1} + \mathbf{b}_{z}) \\
  \mathbf{\rlvn}_{t}&=\sigma(\Weight_{ir}\Obs_t + \Weight_{hr}\Hid_{t-1} + \mathbf{b}_{r}) \\
  \mathbf{n}_{t} &=\sigma(\Weight_{ih} \Obs_t + \mathbf{b}_{ih} + \mathbf{\cont}_{t-1}\odot(\Weight_{hh} \Hid_{t-1}+ \mathbf{b}_{hh})) \\
 \Hid_{t} &= (1 - \mathbf{\rlvn}_t) \odot\mathbf{n}_{t} + \mathbf{\rlvn}_t \odot\Hid_{t-1},
\end{align}
In the backward pass, two cases can be considered, namely unit-wise BRU (UBRU):
\begin{equation}
\Hid'_{t-1} = \Hid_t'\odot\mathbf{\cont}_t + \Hid_{t-1}\odot(1-\mathbf{\cont}_{t})
\end{equation}
and layer-wise BRU (LBRU):
\begin{align}
 \mathbf{\back}_{t}&=\sigma(\Weight_{is}\Obs_t +\mathbf{b}_{is} + \Weight_{hs}\Hid_{t-1} + \mathbf{b}_{hs}) \\
\Hid'_{t-1} &=(\Weight_{hhb}\Hid_t'+\mathbf{b}_{hhb})\odot\mathbf{\back}_t + \Hid_{t-1}\odot(1-\mathbf{\back}_{t}).
\end{align}
Note that in the above, we retain the ad-hoc gate recurrence as we find that it performs marginally better than not doing so.  However, there is currently no probabilistic reason to do so.  We set this matter aside for the future.  With reference to section \ref{sec:forget}, in defining the gates as vectors, we are assuming one gate per feature; this is usual in LSTM and GRU, but not a constraint.
}


\section{Experiments}
\label{sec:experiments}

We present evaluations of the techniques described thus far on automatic speech
recognition (ASR) tasks.  Recurrent networks are particularly suited to ASR as
there is an explicit time dimension and well known context dependency.
Reciprocally, ASR is a difficult task that has driven recent advances in deep
learning \cite{Graves2006, Seide2011, Xiong2017, Hadian2018}.

\subsection{Hypotheses}

In running experiments, we are testing the Bayesian recurrent unit (BRU)
derived in the previous three sections.  This raises two explicit hypotheses:
\begin{enumerate}
\item We would expect the incorporation of a backward pass to improve upon the
  performance of a (forward-only) GRU.
\item We would expect the LBRU to approach the performance of a conventional
  GRU-based BiRNN architecture.  It has the same contextual knowledge, but does
  not have higher representational capability.  If it falls short of a BiRNN
  architecture then either the approximations in the derivation are not valid,
  or the BiRNN is taking advantage of temporal asymmetry in the data.
\end{enumerate}
This is all dependent upon the number of parameters: A BiRNN has roughly twice
as many parameters as a Bayesian RNN with a backward pass.

\subsection{Corpora and method}
Detailed statistics of the corpora considered in this work are summarised in
table \ref{tab:data}.
\begin{table}[ht]
  \centering
  \caption{Statistics of datasets used in this work: speakers and sentences are
    counts, the amounts of speech data for training and evaluation sets are in
    hours.}
\label{tab:data}
\begin{tabular}{|c||c c c c|}
\hline
Dataset & Speakers & Sentences & Train & Eval \\
\hline\hline
TIMIT & 462 & 3696 & 5 & 0.16 \\
WSJ  & 283 & 37416 & 81.3 & 0.7 \\
AMI-IHM & 10487& 98397 & 70.3 & 8.6\\
\hline
\end{tabular}
\end{table}

A first set experiments with the TIMIT corpus \cite{Garofolo1993} was performed to test the proposed model for a phoneme recognition task. We used the standard 462-speaker training set and removed all SA records, since they may bias the results. A separate development set of 50 speakers was used for tuning all meta-parameters including the learning schedule and multiple learning rates. Results are reported using the 24-speaker core test set, which has no overlap with the development set. \rev{Following the implementation of \cite{Ravanelli2018,Ravanelli2019}}, all the recurrent networks tested on this dataset have 5 layers, each consisting 550 units in each direction and use 40 fMLLR features (extracted based on the Kaldi recipe) as the input.

The second set of experiments was carried out on the Wall Street Journal (WSJ) speech corpus to gauge the suitability of the proposed model for large vocabulary speech recognition. We used the standard configuration si284 dataset for training, dev93 for tuning hyper-parameters, and eval92 for evaluation. All the tested recurrent networks have 3 layers, each consisting of 320 units in each direction. We used 40 fMLLR features as input for speaker adaptation.

The TIMIT and WSJ datasets yield results with modest statistical significance.  In order to yield more persuasive significance, a set of experiments was also conducted on the AMI corpus \cite{Carletta2005} with the data recorded through individual headset microphones (IHM). The AMI corpus contains recordings of spontaneous conversations in meeting scenarios, with 70 hours of training data, 9 hours of development, and 8 hours of test data. All the tested recurrent networks have 3 layers, each consisting of 512 units in each direction and use 40 fMLLR features as the input.

The neural networks were trained to predict context-dependent phone targets. The labels were derived by performing a forced alignment procedure on the training set using GMM-HMM, as in the standard recipe of Kaldi\footnote{\url{http://kaldi-asr.org/}} \cite{Povey2011b}. During testing, the posterior probabilities generated for each frame by the neural networks are normalised by their priors, then processed by an HMM-based decoder, which estimates the sequence of words by integrating the acoustic, lexicon and language model information. The neural networks of the ASR system were implemented in PyTorch\footnote{\url{https://pytorch.org/}}, including, crucially, the gradient calculation; they were coupled with the Kaldi decoder \cite{Povey2011b} to form a context-dependent RNN-HMM speech recogniser.

\subsection{Training details}
The network architecture adopted for the experiments contains multiple recurrent layers, which are stacked together prior to the final softmax context-dependent (senon) classifier. If the networks are bidirectional, the forward hidden states and the backward hidden states at each layer are concatenated before feeding to the next layer.  A dropout rate of $0.2$ was used for regularisation. Moreover, batch normalization \cite{Ioffe2015} was adopted on each layer to accelerate the training. The optimization was performed using the Adaptive Moment Estimation (Adam) algorithm \cite{Kingma2014} running for 24 epochs with $\beta_1=0.9$, $\beta_2=0.999$, $\epsilon=10^{-8}$. The performance on the cross validation set was monitored after each epoch, while the learning rate was halved when the performance improvement dropped below a certain threshold ($0.001$).

\subsection{Phoneme recognition performance on TIMIT}
\begin{figure}[htb]
  \centering
  \includegraphics[width=0.9\columnwidth]{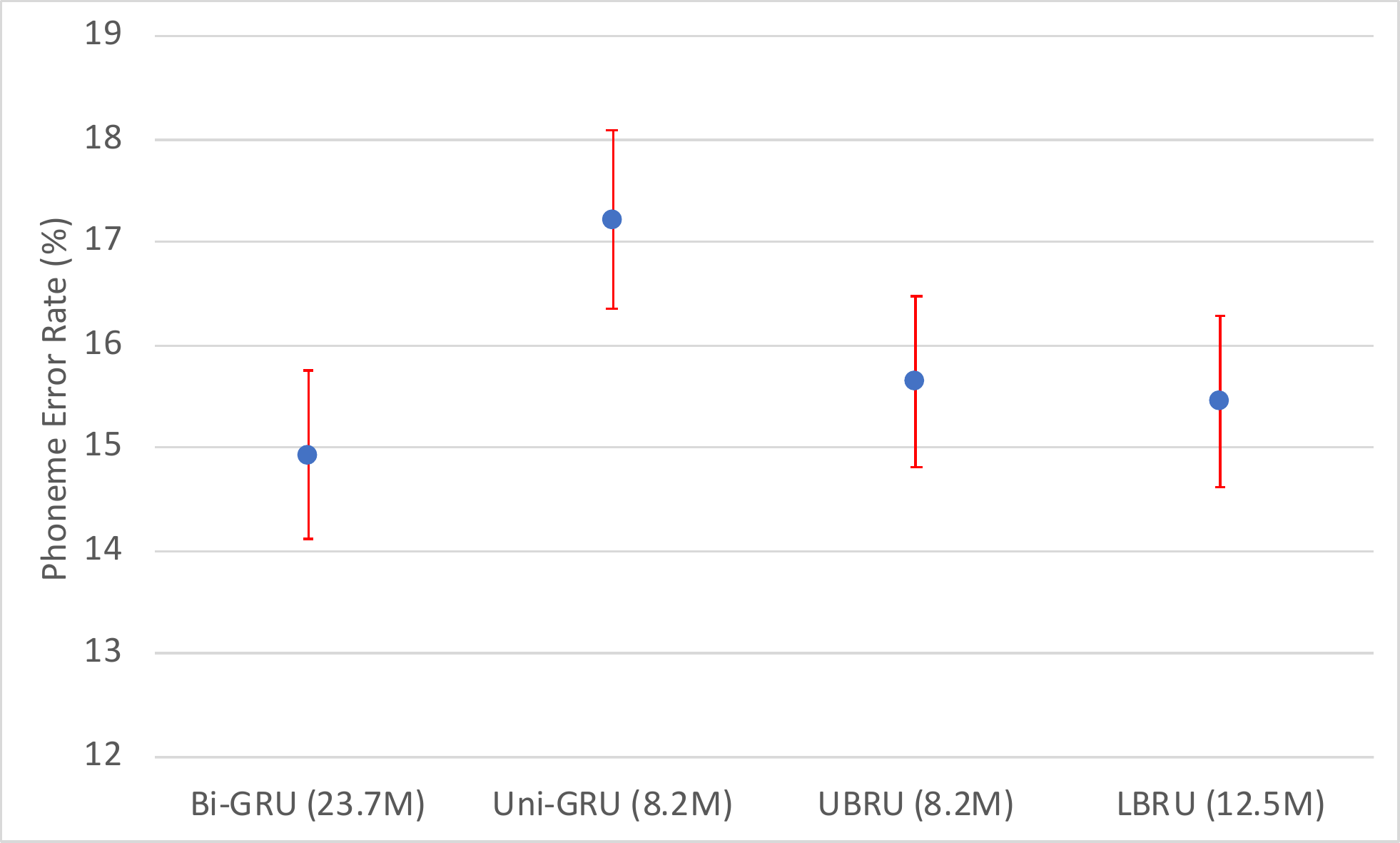}
  \caption{Phoneme Error Rate (\%) on TIMIT for various RNN architectures.}
  \label{fig:timit1}
\end{figure}
In order to confirm the suitability of the proposed model for acoustic modeling, TIMIT was first considered to reduce the linguistic effects (such as lexicon and language model) on the performance evaluation. \rev{The state of the art for this task is probably that of Ravanelli et al.~\cite{Ravanelli2018}, with a phone error rate (PER) of 14.9\%.  We duplicate the architecture of those authors and aim for a similar figure.}  We performed the comparison with GRU as shown in Fig. \ref{fig:timit1}. The error bars indicate equal-tailed 95\% credible interval for a beta assumption for the error rate. The numbers in the parentheses indicate the number of parameters each model contains. It is clear that the unidirectional GRU (Uni-GRU) is significantly worse than bidirectional GRU (Bi-GRU) as the credible intervals do not overlap. By contrast, the unit-wise BRU (UBRU) yields much better performance compared to Uni-GRU with exactly the same model size, and the layer-wise BRU (LBRU) is slightly better than UGRU, yielding similar performance to Bi-GRU.

Since the test set in TIMIT is quite small, we also performed a matched-pair t-test between Uni-GRU and UBRU, the test statistic being the utterance-wise difference in word-level errors normalised by the reference length. This yields $p<0.001$, showing that the UBRU is significantly better. This confirms our first hypothesis that the incorporation of a backward pass can improve upon the performance of a unidirectional GRU. The t-test between Bi-GRU and LBRU yields $p = 0.230$, which implies there is no significant difference between the two systems. The two comparisons together show that our proposed model can achieve performance indistinguishable from the Bi-GRU, without the explicit extra backward recurrence.

\begin{figure}[htb]
  \centering
  \includegraphics[width=0.9\columnwidth]{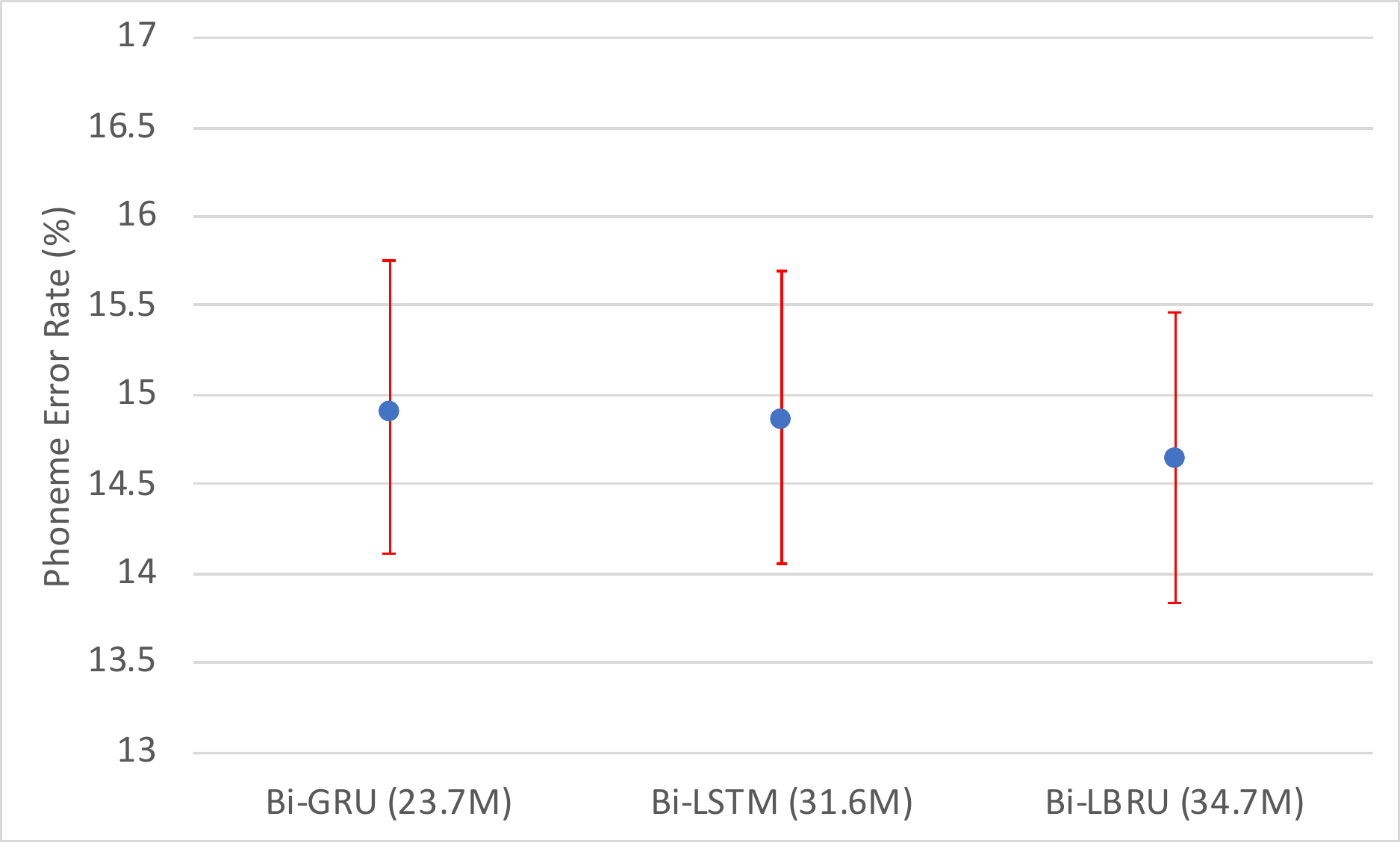}
  \caption{Phoneme Error Rate (\%) on TIMIT for various RNN architectures.}
  \label{fig:timit2}
\end{figure}

Although the difference between Bi-GRU and LBRU is not significant, the latter one is slightly worse. This can be explained by our second hypothesis. Physiological filters are known to have asymmetric impulse responses \cite{Honnet2017}. This is one explanation for the large improvement arising from doubling up the Uni-GRU to explicitly modelling the backward recursion. However, the proposed BRU does not have the explicit extra backward recurrence of the BiRNN architectures. Therefore, we further doubled up the LBRU to be explicitly bidirectional and compared it with Bi-GRU and Bi-LSTM, as shown in Fig. \ref{fig:timit2}. Similarly, we plot the error bars and the sizes of the models; this shows that GRU and LSTM perform almost the same while the Bi-LBRU seems to be slightly better with a few more parameters, although the difference is insignificant from the t-test ($p=0.43$). Our hypothesis is that BRU has a stronger modelling ability in each of the directions because the prediction is always conditioned on the whole sequence due to the implicit backward recursion. \rev{We note that the average PER of 14.6\% obtained with Bi-LBRU outperforms the state of the art 14.9\% of \cite{Ravanelli2018} on the TIMIT test-set, although it is well within the 95\% confidence bounds.}

%

\subsection{Speech recognition performance on WSJ}

\begin{figure}[htb]
  \centering
  \includegraphics[width=0.9\columnwidth]{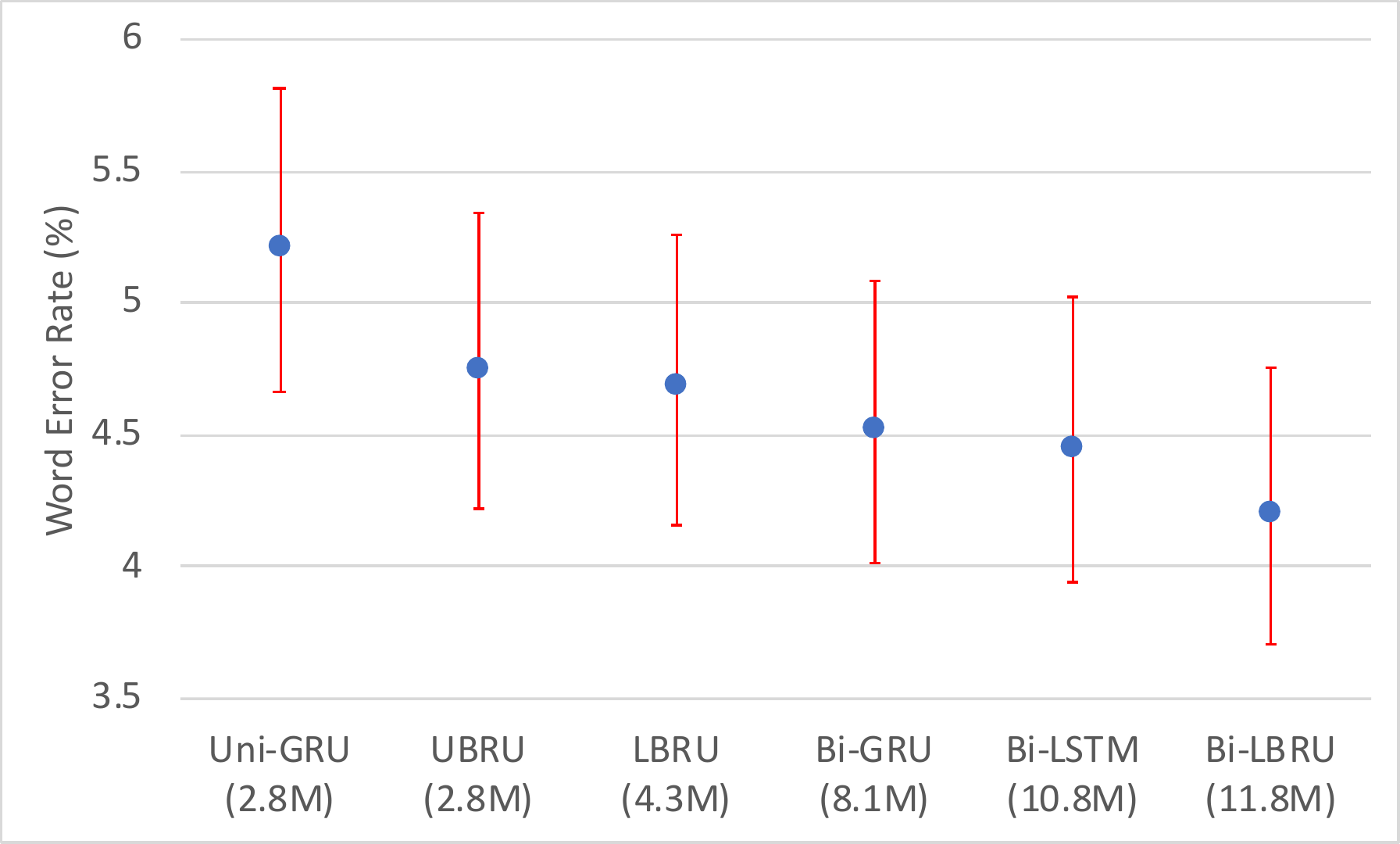}
  \caption{Word Error Rate (\%) on WSJ for various RNN architectures.}
 \label{fig:wsj}
\end{figure}

Since TIMIT is too small to yield significant comparisons, in this sub-section, we evaluate the RNNs on WSJ, a large vocabulary continuous speech recognition task.
Following the TIMIT case, we plot the word error rate (WER) in Fig. \ref{fig:wsj}, together with the corresponding error bars and model sizes. These results exhibit a similar trend to that observed on TIMIT. Both UBRU and LBRU outperform the Uni-GRU ($p=0.19$ from the t-test). LBRU is slightly better than UBRU and it yields very similar performance to that of Bi-GRU ($p=0.21$ from the t-test). The Bi-LBRU still performs slightly better than Bi-GRU and Bi-LSTM. Again, the differences are not significant owing to the fact that the test set of WSJ is still quite small.  \rev{Overall, the results are comparable with the baselines reported in the Kaldi software; for instance, 4.27\% using a Bi-LSTM and i-vectors.}

\subsection{Speech recognition performance on AMI}
\begin{figure}[htb]
  \centering
  \includegraphics[width=0.9\columnwidth]{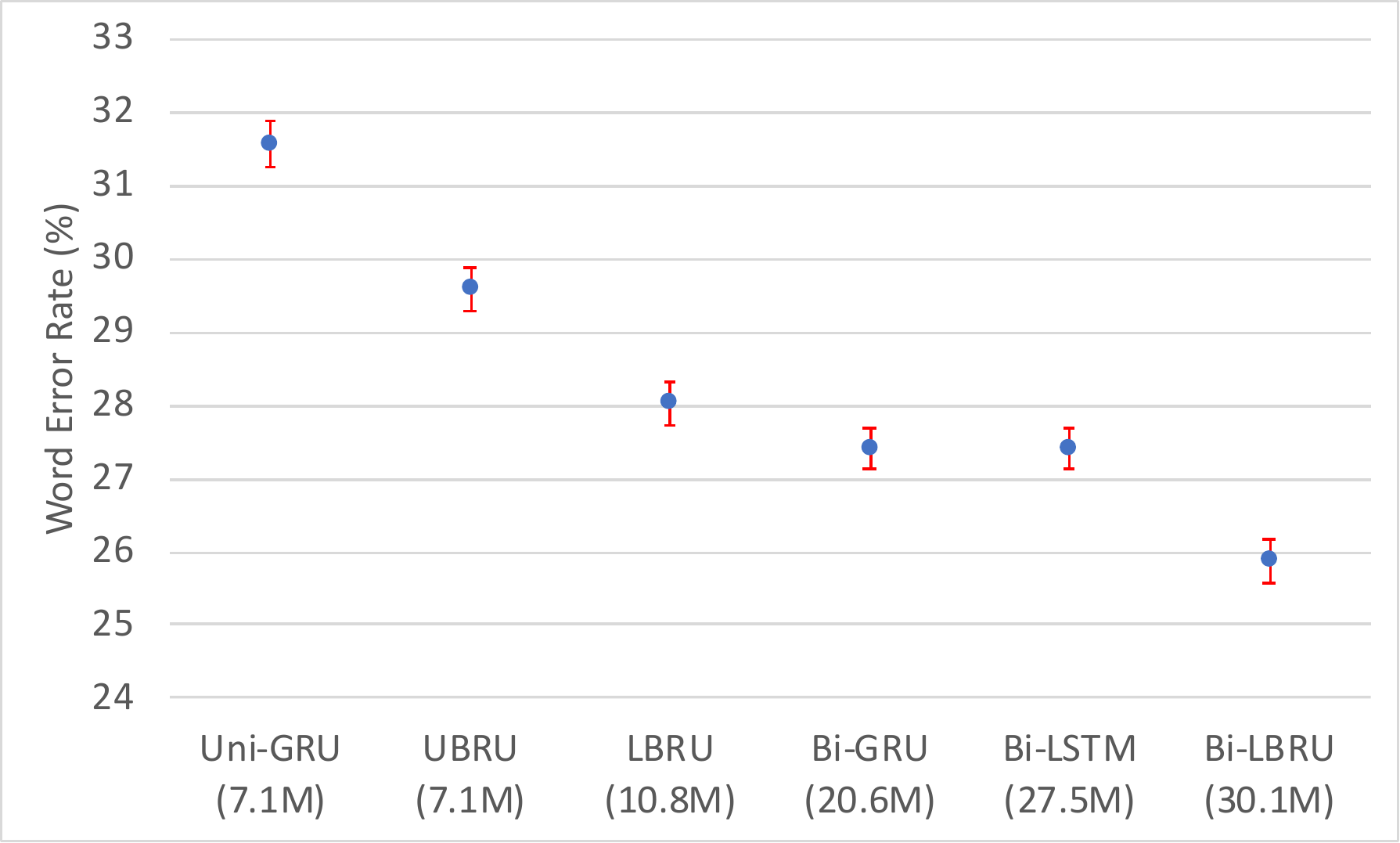}
  \caption{Word Error Rate (\%) on AMI for various RNN architectures.}
 \label{fig:ami}
\end{figure}
Owing to the small test set of WSJ, in this sub-section we conduct the evaluation on AMI, which is a more challenging task with a much larger test set.
AMI is more challenging as the data is recorded in meetings, capturing natural spontaneous conversations between participants who play different roles in the meeting. Overlapping speech segments appear in both training and testing.  \rev{State of the art results on AMI tend to be for complicated systems with elements of speaker and environment adaptation, e.g., Kanda et al \cite{Kanda2018} report a WER of 17.84\%.  Rather than aim to duplicate such results, we simply aim for a self-consistent comparison of techniques; our results are in the same range as the 26.8\% of Dighe et al \cite{Dighe2018}.}

Fig. \ref{fig:ami} summarises the results obtained on AMI. These results show the same trend as previous experiments, but also exhibit more significant differences. Both UBRU and LBRU significantly outperform Uni-GRU while LBRU is also significantly better than UBRU ($p<0.001$ from the t-test), showing that the layer-wise backward recursion is able to capture richer characteristics in the backward transition. Comparison between LBRU and Bi-GRU shows that LBRU can achieve similar performance without an extra explicit backward network. Bi-LSTM does not have any advantages over Bi-GRU, although it contains one more gate and, therefore, more parameters. However, if we double up the LBRU to be explicitly bidirectional, the model yields significantly better performance than both Bi-GRU and Bi-LSTM ($p<0.001$ from the t-test). This confirms the hypothesis that BRU has a stronger unidirectional modelling ability and explicit bidirectional modelling can help capture the asymmetric characteristics in physiological filters.


\section{Conclusion}

Given a probabilistic interpretation of common neural network components, it is
possible to derive recurrent components in the same spirit.  \rev{Such components have two advantages:
\begin{enumerate}
\item The architecture of the recursion is dictated by the probabilistic formulation, removing otherwise ad-hoc choices.
\item They naturally support a backward recursion of the type used in Kalman smoothers and the forward-backward algorithm of the HMM.
\end{enumerate}}

Unit-wise recursions follow analytically, but are found to lead to
instabilities.  Approximations lead to stable layer-wise recursions.
Nevertheless, useful backward recursions can be derived for both cases.  The
resulting Bayesian recurrent unit (BRU) can be configured with a probabilistic
input gate, being directly comparable to a common GRU.

Evaluation on simple and on state of the art speech recognition tasks shows
that:
\begin{enumerate}
\item Even the unit-wise backward recursion can out-perform a standard GRU.
\item A more involved layer-wise backward recursion can approach the
  performance of a bidirectional GRU.  This shows that the approximations in
  the derivations are reasonable.
\end{enumerate}
Further, an explicit bidirectional BRU can out-perform a state of the art
bidirectional GRU.

There are some ad-hoc methods in our approach: the gate recurrences are retained for performance; some approximations may be
better formulated.  These remain matters for future research.  Nevertheless, we have shown that recurrence in neural
networks can be formulated much more rigorously than conventional wisdom would
hold.  This in turn can lead to significant performance advantages.

\appendices


\ifCLASSOPTIONcompsoc
  \section*{Acknowledgments}
\else
  \section*{Acknowledgment}
\fi

We are grateful to our colleagues Apoorv Vyas and Bastian Schnell for battling through an early manuscript and providing comments that rendered it more accessible.  \rev{We also extend our thanks to the editor and two referees for their insights.}

Part of this work was conducted within the scope of the Research and Innovation
Action SUMMA, which has received funding from the European Union's Horizon 2020
research and innovation programme under grant agreement No 688139.

\ifCLASSOPTIONcaptionsoff
  \newpage
\fi



\bibliographystyle{IEEEtran}
\bibliography{png-strs,png-refs,png-pubs,png-tech}
\end{document}